\documentclass[letterpaper, 10 pt, conference]{ieeeconf}
\IEEEoverridecommandlockouts
\usepackage{amsmath,amssymb,amsfonts}
\usepackage{algorithmic}
\usepackage{graphicx}
\usepackage{textcomp}
\usepackage{xcolor}
\usepackage{units}
\usepackage{booktabs}
\usepackage{multirow}
\usepackage{subfig}
\usepackage{stfloats}
\usepackage{cite}
\usepackage{tikz}

\newcommand\copyrighttext{%
  \footnotesize \textcopyright 2024 IEEE. Personal use of this material is permitted.
  Permission from IEEE must be obtained for all other uses, in any current or future 
  media, including reprinting/republishing this material for advertising or promotional 
  purposes, creating new collective works, for resale or redistribution to servers or 
  lists, or reuse of any copyrighted component of this work in other works. 
  }
\newcommand\copyrightnotice{%
\begin{tikzpicture}[remember picture,overlay]
\node[anchor=south,yshift=10pt] at (current page.south) {\fbox{\parbox{\dimexpr\textwidth-\fboxsep-\fboxrule\relax}{\copyrighttext}}};
\end{tikzpicture}%
}

\title{\LARGE \bf
An Analysis of Driver-Initiated Takeovers during Assisted Driving\\and their Effect on Driver Satisfaction
}

\author{Robin Schwager$^{1}$, Michael Grimm$^{1}$, Xin Liu$^{2}$, Lukas Ewecker$^{1}$,\\Tim Brühl$^{1}$, Tin Stribor Sohn$^{1}$ and Sören Hohmann$^{3}$
\thanks{$^{1}$Dr. Ing. h.c. F. Porsche AG, Weissach, Germany.
        {\tt\small (robin.schwager1,michael.grimm3,lukas.ewecker, tim.bruehl,tin\_stribor.sohn) @porsche.de}}%
\thanks{$^{2}$Dresden University of Technology, Dresden, Germany.
        {\tt\small xin.liu2@mailbox.tu-dresden.de}}%
\thanks{$^{3}$Karlsruhe Institute of Technology, Karlsruhe, Germany.
        {\tt\small soeren.hohmann@kit.edu}}%
}

\begin{document}

\maketitle

\copyrightnotice

\bibliographystyle{IEEEtran}

\thispagestyle{empty}
\pagestyle{empty}

\begin{abstract}

During the use of Advanced Driver Assistance Systems (ADAS), drivers can intervene in the active function and take back control due to various reasons.
However, the specific reasons for driver-initiated takeovers in naturalistic driving are still not well understood.
In order to get more information on the reasons behind these takeovers, a test group study was conducted.
There, 17 participants used a predictive longitudinal driving function for their daily commutes and annotated the reasons for their takeovers during active function use.
In this paper, the recorded takeovers are analyzed and the different reasons for them are highlighted.
The results show that the reasons can be divided into three main categories.
The most common category consists of takeovers which aim to adjust the behavior of the ADAS within its Operational Design Domain (ODD) in order to better match the drivers' personal preferences.
Other reasons include takeovers due to leaving the ADAS's ODD and corrections of incorrect sensing state information.
Using the questionnaire results of the test group study, it was found that the number and frequency of takeovers especially within the ADAS's ODD have a significant negative impact on driver satisfaction.
Therefore, the driver satisfaction with the ADAS could be increased by adapting its behavior to the drivers' wishes and thereby lowering the number of takeovers within the ODD.
The information contained in the takeover behavior of the drivers could be used as feedback for the ADAS.
Finally, it is shown that there are considerable differences in the takeover behavior of different drivers, which shows a need for ADAS individualization.
\end{abstract}

\section{Introduction}

Vehicles equipped with Advanced Driver Assistance Systems (ADAS) are becoming increasingly popular in the automotive market.
Especially level 1 or 2 functions, as defined by the Society of Automotive Engineers (SAE) \cite{SAE2021}, are commonly found in state-of-the-art vehicles.
When using functions like these, drivers hand over parts of vehicle control to the ADAS, which puts the vehicle in a state of mixed control.
During this state, both driver and ADAS can decide to stop the automation and hand full control back to the driver.
This is commonly referred to as a takeover of control.
System-initiated takeover requests occur most of the time when the vehicle detects that it leaves the Operational Design Domain (ODD) of the ADAS and the driver is required to resume control to handle the situation \cite{Moralez2020}.
However, especially at low-level automation, driver-initiated takeovers are much more common than system-initiated takeovers \cite{Morando2020, Gershon2021}.
Drivers are required to continuously monitor the driving environment during the use of SAE level 1 and 2 functions and have to be prepared to take over the vehicle control task at any time.
Leaving the ODD of the ADAS is therefore one main reason for driver-initiated takeovers since often low-level ADAS do not detect scenarios in which they are outside of their ODD.
E.g., an Adaptive Cruise Control (ACC) will only react to leading vehicles but not to other road users in give-way situations.
The driver must then recognize this scenario and take back control of the vehicle manually.
Sometimes, the ADAS might also make erroneous decisions due to incorrect input information from the vehicle's sensors.
Therefore, drivers must also be prepared to correct any potential mistakes of the ADAS.

However, being forced to observe the ADAS behavior so closely means that the driver is fully aware of the driving function's behavior, causing them to notice differences in the function's driving style and their own preferred style of driving.
If the differences are strong enough, the driver might intervene, adjusting the ADAS's behavior accordingly, and then hand back control to the system.
This is the third reason for driver-initiated takeovers: a takeover to adjust the system's behavior based on the driver's personal preferences.
These interventions in the active function could be used as feedback for ADAS optimization and individualization.
The aim of this work is to analyze the different types of driver-initiated takeovers that occur during active ADAS use and to evaluate their potential as a source for ADAS optimization based on the results of a test group study conducted on German roads.

An "intervention" in the context of this work is a short-term driver-initiated takeover of vehicle control.
After the driver intervenes to correct the function behavior, they give control back to the system.
Another type of intervention would be an adjustment of the system behavior without directly taking over control, like a set speed increase.

\section{Related work}

\subsection{Driver-Initiated Takeovers}

Although \cite{Morando2020, Gershon2021} state that driver-initiated takeovers are more common in the real world, most studies focus on system-initiated takeovers.
There, the main topic is often to analyze the driver's readiness for takeover and their takeover performance, especially when the driver is immersed in secondary tasks during driving \cite{Moralez2020, Rydstrom2023, Gershon2023, Braunagel2017}.

However, driver-initiated takeovers have only been analyzed by a few authors until now.
References \cite{Yang2023, Gershon2021} both work on categorizing driver-initiated takeovers during naturalistic driving.
For this, they use the Massachusetts Institute of Technology (MIT) Advanced Vehicle Technology (AVT) study dataset \cite{Fridman2019}, which is a publically available dataset containing naturalistic real-world driving data of human drivers using SAE level 1 and 2 driving functions.
The dataset contains video material of three to six cameras, raw vehicle bus data, Global Positioning System (GPS) data, and IMU data from \unit[823]{km} of driving by 122 drivers in 29 vehicles.
The two SAE level 2 ADAS systems used for takeover analyses in this dataset are the Tesla Auto Pilot (AP) and Cadillac Super Cruise (SC).

Using this dataset, \cite{Yang2023} extracted 739 driver-initiated takeovers from the data of eight Cadillac drivers using SC and eight Tesla drivers using AP and enriched them with context information.
The extracted driver-initiated takeovers were then clustered using agglomerative hierarchical clustering.
The clustering found four main categories of driver-initiated takeovers for both functions, called \textit{normal takeovers}, \textit{braking takeovers}, \textit{accelerating takeovers}, and \textit{evasive-maneuver takeovers}.
Due to the limited number of features extracted from the data, only a rough categorization can be provided but it gives a first overview of potential main categories of driver-initiated takeovers.

Reference \cite{Gershon2021} also used the MIT AVT study dataset to extract and manually annotate 428 complete automation disengagements, either driver- or system-initiated, from 14 Cadillac SC drivers.
They classified the reasons behind the driver-initiated takeovers as \textit{strategic}, \textit{maneuver}, or \textit{control}, based on the hierarchical model of driver behavior \cite{Michon1985}.
They then analyzed specific characteristics of the takeover types like, e.g., their kinematics, duration, and whether drivers re-engaged the automation afterwards.

However, due to missing ground truth data about the drivers' explicit reasons to take over the system in each case, analyses on the different categories of driver-initiated takeovers in the MIT AVT study dataset are limited to feature extractions and rough clusterings like these.

\subsection{ADAS Personalization and Satisfaction}

Several studies have shown that driving styles differ across multiple drivers \cite{TaubmanBenAri2004, Sagberg2015}.
Therefore, many works focus on creating driver models by observing the driver's behavior during manual drives.
These models can then be used to parameterize automated driving functions to mimic the individual driving styles \cite{Hasenjager2020, Kuderer2015, Wang2014, Lin2014}.

Other studies were conducted to investigate whether drivers even prefer their own driving style for automated driving \cite{Hasenjager2020}.
References \cite{Basu2017, Yusof2016} argue that both aggressive and defensive drivers prefer a defensive automated driving style.
On the other hand, \cite{Griesche2016, Ma2021} found that drivers prefer an automated driving style closer to their own, i.e., that aggressive drivers do in fact prefer an aggressive automated driving style.
Some studies point out that this may also be related to other factors, e.g., the driver's age \cite{Hartwich2018, Scherer2015}.
The designs of these studies often vary significantly and some studies focus on scenarios that are absent in other studies, making comparability difficult.
Also, most of these studies are not conducted in real-world driving scenarios with the participant in the driver seat but rather use driving simulators \cite{Basu2017, Griesche2016} or seating the study participant in the passenger seat or back seat \cite{Vasile2023, Yusof2016}.
But all of these studies agree that there are benefits to adjusting and individualizing the ADAS driving style.

For the optimization and individualization of driving functions, some works propose a continuous, iterative process \cite{Hasenjager2020, Chen2017}.
This could provide benefits compared to the previous approaches where manual driving behavior is observed and the driver model is derived from that alone.
Instead, continuous feedback by the driver during the use of an ADAS could be utilized for iterative optimization.
The field of interactive imitation learning also builds up on the promise that human intervention and correction in an active system can be used as feedback for optimization \cite{Celemin2022}.
It is commonly accepted that drivers who are unsatisfied with an ADAS will intervene and take over control of the system to correct its behavior.
Some studies even use the driver's interventions as an indication of dissatisfaction \cite{Ma2021, Lee2021}.
However, to the author's best knowledge, no study has proven the correlation between drivers' interventions and their satisfaction with an ADAS yet.

\section{Assisted Driving Function}

The assisted driving function used for this study is an in-production Predictive Longitudinal Driving Function (PLDF) with the SAE level 1.
The PLDF uses its input information, mainly map information about legal speeds and road topography, to calculate a pleasant speed profile for the road ahead in a specified time window.
It also encompasses ACC for vehicle following which overrides the calculated speed profile if necessary.
Lateral control is not part of the PLDF, meaning the steering is still done manually by the driver.
Therefore, the scope of the system is the following:

\begin{itemize}
\item Compliance to the legal speed limit
\item Speed reduction before and in curves, turns, roundabouts and give-way situations
\item ACC target vehicle following
\end{itemize}

The legal speed limits from the map information are reconfirmed by a camera-based traffic sign recognition system.
For curves, turns, and roundabouts, the aforementioned map information is used to calculate pleasant speed profiles.
For the ACC, a radar and camera fusion is used for target vehicle detection.
Additionally, the PLDF accepts manual driver inputs.
Using a lever, the driver can activate and deactivate the PLDF and adjust the chosen set speed and ACC time gap.
Drivers can also override the PLDF behavior by accelerating above or decelerating below the speed chosen by the system by using the gas and brake pedal.

Besides following the leading vehicle, all interactions with traffic fall outside of the PLDF's ODD.
Therefore, the driver must intervene in the active system when such a traffic scenario occurs.
These scenarios mainly include stopping or decelerating at red traffic lights and reacting to other traffic participants in give-way situations.
After a prolonged standstill, the driver must also reactivate the function manually.
Besides these scenarios, the PLDF is designed to handle full longitudinal control over a whole drive without manual interventions by the driver.
However, there are scenarios where the PLDF works within its ODD, and drivers still correct its behavior by changing the set speed, pressing the gas or brake pedal, or deactivating the system.
These corrections can either be a result of incorrect input information from the map or sensing state or of a differing preference by the driver.
In summary, there are three main reasons for a driver to intervene in the active system:

\begin{enumerate}
    \item Takeovers required by system design when leaving the ODD
    \item Corrections of incorrect map or sensor data
    \item Corrections of intended system behavior based on a deviating personal preferences of the driver
\end{enumerate}

These categories will be investigated more in detail in this work.

\section{Test Group Study}

\subsection{Scope}
In contrast to most other studies on driver satisfaction with automated or assisted driving functions, this study focuses on recording naturalistic driving data as close as possible to real customer behavior.
The goal is to generate a dataset that represents real driving behavior with the PLDF and the drivers' interventions in said function.
This dataset allows for the evaluation of whether these interventions affect driver satisfaction negatively.
If so, the dataset can be used to analyze the occurring interventions and deduce potential optimizations of the system behavior.

\subsection{Prerequisites}

Firstly, the participants should already be sufficiently familiar with the PLDF.
Pre-tests have shown that drivers unfamiliar with the system tend to use it more cautiously, needing to become familiar with the PLDF before correcting behavior that deviates from their preferred driving style.

Secondly, the drivers should be familiar with the driven route, since pre-tests also showed that drivers drive more cautiously on unknown routes compared to routes they knew well.

Third, due to being set in the real world, drives on the same route can differ significantly depending on the time of day, weather, traffic volume, driver mood, and more.
Therefore, multiple drives by each participant provide a better representation of all relevant scenarios and make the results more robust.

Fourth, due to the PLDF's design, rural routes are preferred over freeway or urban driving.
On freeways, almost only the ACC component of the PLDF comes into play, while in urban scenarios, frequent interactions with the surrounding traffic might reduce the system uptime significantly.

Finally, related work in the sector of driver-initiated takeover categorization has shown that deducing the reason for a takeover just from the recorded vehicle data can be difficult and ambiguous \cite{Yang2023, Gershon2021}.
Therefore, it was decided that the participants should annotate each intervention they performed and state their reasons for correcting the system.
This way, ground truth information for each takeover was gathered for in-depth analysis.

\subsection{Setup}
The test group study was conducted in southwestern Germany.
Participants were recruited internally at Dr. Ing. h.c. F. Porsche AG based on their experience with the PLDF and the suitability of their commuter route to fit the abovementioned requirements.
Two vehicles were used for the study, one Porsche Cayenne from 2023 and one Porsche 911 from 2022.
Both test vehicles were equipped with the PLDF, a logger for bus data, and a smartphone for voice annotations.

In total, 26 participants took part in the study.
Due to multiple reasons like annotation quality and problems with the data loggers, the data of some participants had to be filtered out, resulting in a final dataset of 17 participants.

\subsection{Procedure} \label{subsec:procedure}
In the beginning, each participant filled out the first half of the questionnaire inquiring about demographic data like age, general driving experience, and experience with assisted or automated driving functions.
Then, the task of the study was explained to the participants.

They should use the test vehicle for one week for their daily commute to and from work while using the PLDF.
During these drives, participants should use the PLDF as much as possible by avoiding longer sections of manual driving.
But they were also told to freely take over the vehicle control whenever they felt the need to.
After each takeover, they should hand back control to the PLDF and annotate the reason for their intervention using the smartphone mounted in the car.
Each voice annotation should contain specific information about different aspects of the intervention:

\begin{enumerate}
    \item \textit{Driver input}
    \item \textit{Situation}
    \item \textit{Reason}
    \item \textit{Desired behavior}
\end{enumerate}

By the \textit{driver input}, the action they performed is meant, i.e., whether they used the gas pedal, brake pedal, or the control lever for cancelation or set speed adjustments.
Examples for the \textit{situation} would be a curve, when approaching a turn, when entering a roundabout, or on a straight road.
The \textit{reason} describes why the driver intervened.
This can mostly be broken down to traffic-related interventions, incorrect input data, or a deviating personal preference.
If the \textit{reason} was such a personal preference, the \textit{desired behavior} should explain, what the driver wanted instead, e.g., a higher speed or a later deceleration.

It was also explained to the drivers that they should always drive the same route to and from work without any additional stops.
This way, the driver behavior on the same routes traveled multiple times can be better compared.
After receiving the instructions, a short test drive was conducted with each participant, in order to verify whether they understood the instructions and to check whether any questions were still open.
Then, the vehicle was handed over to the study participant.

After the week was over, the participants handed back the vehicle and answered the second half of the questionnaire.
This time, they were asked about their satisfaction with the function using a 5-point Likert scale ranging from "I strongly disagree" to "I strongly agree".
The statements that the participants should judge were:

\begin{itemize}
    \item \textit{I was satisfied with the function in general.}
    \item \textit{A reduction of the needed interventions would increase my satisfaction.}
    \item \textit{I was satisfied with the function outside of the scenarios where I intervened.}
\end{itemize}

\begin{figure*}[b]
\subfloat[Vehicle bus signals and timestamp of the driver's voice annotation plotted over time.]{
\includegraphics[width=0.5\paperwidth]{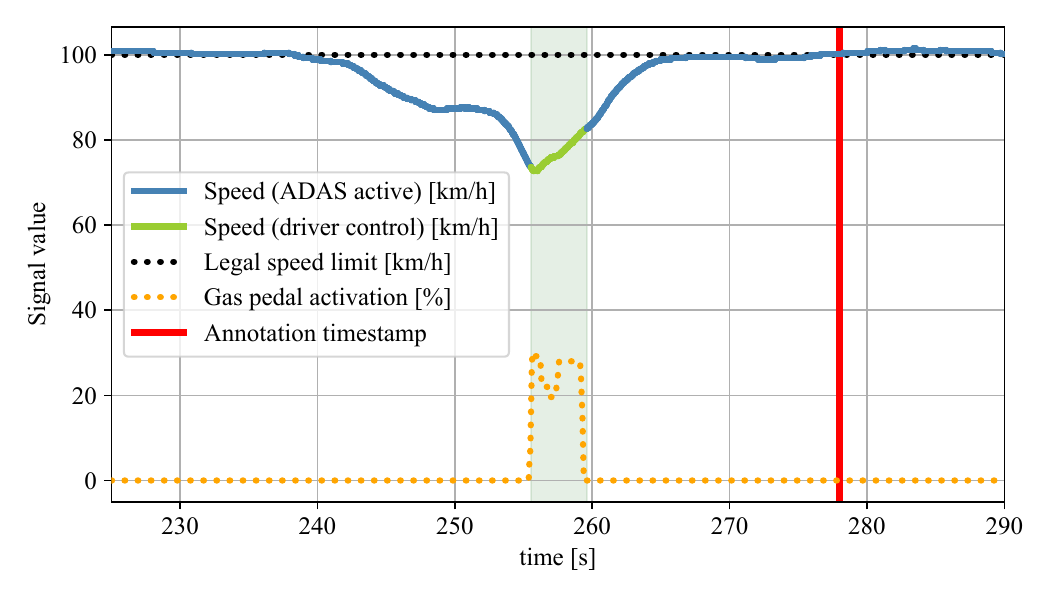}\label{fig:intervention_annotation_example_vt}}
\subfloat[Vehicle GPS locations on a map view.]{
\includegraphics[width=0.32\paperwidth]{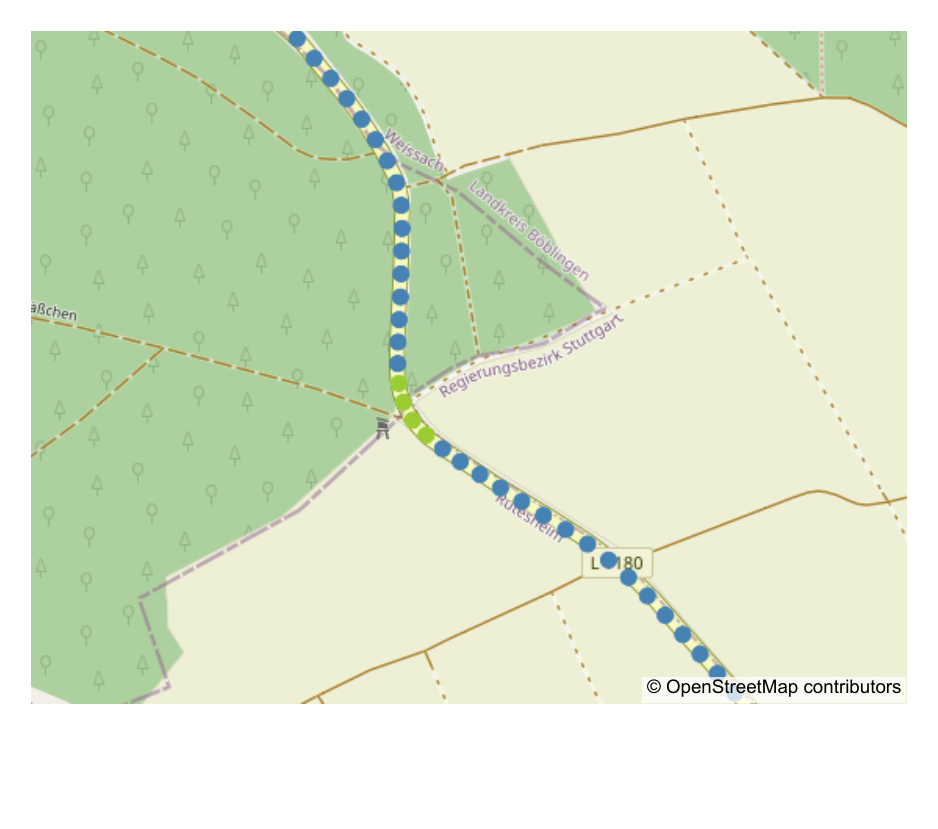}\label{fig:intervention_annotation_example_map}}
\caption{Example intervention on a curvy road. The driver temporarily takes over control by pressing the gas pedal.}
\label{fig:intervention_annotation_example}
\end{figure*}

\subsection{Dataset Creation}

The dataset created in this study consists of the in-vehicle bus communication data and the voice annotations created by the drivers.
After the data was recorded, it had to be preprocessed before further analyses could be carried out.
This includes data cleaning like filtering out non-commuter drives and translating the voice annotations to machine-readable labels.
While the data cleaning could be partially automated, the labels for each intervention had to be annotated manually.
For this manual labeling process, each voice annotation was split into its main four components and the labels were applied accordingly.
These components were explained in subsection \ref{subsec:procedure}.

Interventions with missing or incomplete voice annotations were still annotated if the reasons behind these interventions were obvious from the bus data.
However, cases with unclear driver intentions had to be labeled with the reason "unknown".
This labeling strategy ensures that the number of annotations in the dataset matches the number of interventions.

An example of an intervention is shown in Fig. \ref{fig:intervention_annotation_example}.
There, a participant was driving on a curvy road which can be seen in Figure \ref{fig:intervention_annotation_example_map}.
Due to the high curvature of the road, the driving function decreases the speed below the legal speed to ensure pleasant curve speeds for the driver.
However, the driver is not satisfied with the low speed in the last curve and presses the gas pedal to override the PLDF in the second curve.
This can be seen in Figure \ref{fig:intervention_annotation_example_vt}.
There, selected signals from the recorded bus data are plotted over time for this scenario.
In the intervals in which the PLDF controls the vehicle speed, the speed signal is colored in blue.
As can be seen, when the vehicle decelerates strongly in the second curve, the driver presses the gas pedal, thereby taking control of the vehicle and accelerating it.
The takeover of control is indicated by the green section of the speed signal in Figure \ref{fig:intervention_annotation_example_vt} and the green GPS locations in Figure \ref{fig:intervention_annotation_example_map}.
After the curve is passed, the driver hands back control to the system.
Then, the driver annotates their intervention at the timestamp highlighted with the red vertical line.
The section where the driver took over control of the system is highlighted with a green background color.
During the manual labeling of the dataset, the voice annotation is translated to the previously described label components and its start and end time are set to the green segment.
For this specific scenario, the labels are set as follows:

\begin{itemize}
    \item \textit{Driver input:} Gas pedal
    \item \textit{Situation:} Curve
    \item \textit{Reason:} Personal preference
    \item \textit{Desired behavior:} Higher speed wanted
\end{itemize}

\section{Results}

\subsection{Questionnaire} \label{subsec:questionnaire}
The results for the second half of the questionnaire inquiring about the drivers' satisfaction with the PLDF and their thoughts on the interventions are shown in Table. \ref{tab:questionnaire_results}.
As can be seen, most drivers are relatively satisfied with the PLDF, while some drivers have mixed feelings about it or are even unsatisfied.
It can also be seen that every single driver would subjectively like to reduce the number of interventions by stating that they would be more satisfied with the PLDF if fewer interventions were necessary.
Finally, most drivers agree that the PLDF's behavior outside of the scenarios in which they intervened is mostly satisfactory.
Only three out of 17 drivers gave a mixed answer to this question while only one driver claimed to be also unsatisfied with the PLDF outside of scenarios they intervened in.
This confirms that at least subjectively, reducing the number of interventions is a feasible goal in order to increase driver satisfaction and that most drivers correct function behavior with which they are unhappy.

\begin{table*}[htbp]
\renewcommand{\arraystretch}{1.3}
\caption{Results for the second half of the questionnaire}
\label{tab:questionnaire_results}
\begin{center}
\begin{tabular}{cccccc}
\toprule 
    & \multicolumn{5}{c}{\textbf{Number of participants who chose the option:}}\tabularnewline
    & Strongly disagree & Disagree & Neutral & Agree & Strongly agree\tabularnewline
\midrule
\midrule 
I was satisﬁed with the function in general. & 0 & 2 & 6 & 8 & 1\tabularnewline
\midrule 
A reduction of the needed interventions would increase my satisfaction. & 0 & 0 & 0 & 8 & 9\tabularnewline
\midrule 
I was satisﬁed with the function outside of the scenarios where I intervened. & 0 & 1 & 3 & 9 & 4\tabularnewline
\bottomrule
\end{tabular}    
\end{center}
\end{table*}
    
\subsection{Dataset Overview}

The dimensions of the resulting dataset are shown in Table \ref{tab:dataset_overview}.
As can be seen, although the number of participants is rather low, the amount of gathered data for these is quite extensive.
Especially noteworthy is the high number of interventions in the PLDF which shows that, on average, an intervention occurs every 1.7 minutes of driving.
Out of the \unit[4334]{km} driven, \unit[14.3]{\%} were driven on freeways, while \unit[25.9]{\%} were driven inside towns and cities.
The remaining \unit[59.8]{\%} are comprised of rural roads and other street categories.

\begin{table}[t]
\renewcommand{\arraystretch}{1.3}
\caption{Dataset overview}
\label{tab:dataset_overview}
\begin{center}
\begin{tabular}{cc}
\toprule 
Total number of participants & 17\tabularnewline
\midrule 
Total number of drives & 165\tabularnewline
\midrule 
Total time & \unit[92.8]{h}\tabularnewline
\midrule 
Total distance & \unit[4334]{km}\tabularnewline
\midrule 
Total number of annotations & 3335\tabularnewline
\midrule 
Average number of drives per subject & 9.71\tabularnewline
\midrule 
Average time per drive & \unit[33.7]{min}\tabularnewline
\midrule 
Average distance per drive & \unit[25.6]{km}\tabularnewline
\midrule 
Average number of annotations per drive & 20.2\tabularnewline
\bottomrule
\end{tabular}
\end{center}
\end{table}

\subsection{Types of Interventions}

For the following analysis of the different types of interventions, the detailed labels used in the dataset are grouped and generalized for a better overview and easier understanding.
One problem with the absolute and relative numbers of annotations in the dataset is that the data of participants who drove longer distances and intervened more often in total have a bigger influence on the distribution of labels than the data of drivers who had shorter commutes.
Therefore, the relative distribution of labels over all participants will also be shown.
These rebalanced values represent the average of the relative distribution of annotation labels of all subjects weighted equally.
In this way, the interventions of participants with a higher number of total annotations will not be weighted stronger.

The distribution of the different interventions in the dataset is shown in Table \ref{tab:label_overview}.
There, the different reasons for the interventions are grouped into four clusters.
It can be seen that, for the rebalanced distribution, \unit[53.7]{\%} of interventions in the PLDF are based on personal preferences by the drivers which deviate from the intended function behavior.
The wrong input data from the map, traffic sign detection, and ACC sensing state make up another \unit[12.3]{\%} of interventions.
These two combined describe all interventions that fall within the PLDF's ODD.
This means that if the PLDF both had perfect input data and its behavior would be adjusted to the individual driver wishes, \unit[66.0]{\%} of driver interventions would no longer be necessary.
The system limitations due to traffic make up for \unit[27.7]{\%} of interventions.
These cannot be prevented with the PLDF's current scope but would require the addition of entirely new features instead.

\begin{table*}[t]
\renewcommand{\arraystretch}{1.3}
\caption{Distribution of labels regarding the reason of the intervention}
\label{tab:label_overview}
\begin{center}
\begin{tabular}{ccccc}
\toprule 
\textbf{Cluster} & \textbf{Reason} & \textbf{Absolute} & \textbf{Relative} & \textbf{Rebalanced}\tabularnewline
\midrule
\midrule 
\multirow{5}{*}{Personal preference} & Speed adjustment on straight road & 976 & \unit[29.3]{\%} & \unit[25.6]{\%}\tabularnewline
\cmidrule{2-5} \cmidrule{3-5} \cmidrule{4-5} \cmidrule{5-5} 
    & Speed adjustment before or in curve, turn, or roundabout & 461 & \unit[13.8]{\%} & \unit[14.4]{\%}\tabularnewline
\cmidrule{2-5} \cmidrule{3-5} \cmidrule{4-5} \cmidrule{5-5} 
    & Adjustment of acceleration or deceleration timing onto speed limits & 363 & \unit[10.9]{\%} & \unit[10.0]{\%}\tabularnewline
\cmidrule{2-5} \cmidrule{3-5} \cmidrule{4-5} \cmidrule{5-5} 
    & Adjustment of acceleration strength & 80 & \unit[2.4]{\%} & \unit[2.3]{\%}\tabularnewline
\cmidrule{2-5} \cmidrule{3-5} \cmidrule{4-5} \cmidrule{5-5} 
    & Adjustment of ACC behavior & 45 & \unit[1.4]{\%} & \unit[1.1]{\%}\tabularnewline
\midrule 
\midrule 
System limitations & Traffic interactions & 802 & \unit[24.0]{\%} & \unit[27.7]{\%}\tabularnewline
\midrule 
\midrule 
\multirow{2}{*}{Wrong input data} & Correction of incorrect ACC sensing state & 207 & \unit[6.2]{\%} & \unit[7.0]{\%}\tabularnewline
\cmidrule{2-5} \cmidrule{3-5} \cmidrule{4-5} \cmidrule{5-5} 
    & Correction of incorrect traffic sign recognition or map information & 179 & \unit[5.4]{\%} & \unit[5.3]{\%}\tabularnewline
\midrule 
\midrule 
\multirow{3}{*}{Other} & Unintentional intervention & 77 & \unit[2.3]{\%} & \unit[2.3]{\%}\tabularnewline
\cmidrule{2-5} \cmidrule{3-5} \cmidrule{4-5} \cmidrule{5-5} 
    & Unknown, missing annotation & 93 & \unit[2.8]{\%} & \unit[2.3]{\%}\tabularnewline
\cmidrule{2-5} \cmidrule{3-5} \cmidrule{4-5} \cmidrule{5-5} 
    & Other & 47 & \unit[1.4]{\%} & \unit[1.4]{\%}\tabularnewline
\bottomrule
\end{tabular}
\end{center}
\end{table*}

\subsubsection{Interventions due to Personal Preference}

The most common intervention due to personal preference is the speed adjustment on straight roads with \unit[25.6]{\%} of the rebalanced distribution.
This is done most of the time by increasing the current set speed of the PLDF slightly above the current legal speed using the control lever.
This makes up for \unit[64.1]{\%} of these interventions, while set speed decreases below the legal speed limit make up for \unit[9.0]{\%}.
The acceleration of the vehicle above the legal speed on straight roads via the gas pedal make up for \unit[21.0]{\%}.
The remaining \unit[5.9]{\%} were decreases below the legal speed via the brake.

The second most common intervention due to personal preference is the speed adjustment before or in curves, turns, or roundabouts with \unit[14.4]{\%} of the rebalanced distribution.
There, the PLDF decreases the vehicle speed before reaching the curve, turn, or roundabout, drives through at a speed deemed pleasant, and then accelerates again afterward.
The interventions in this category comprise scenarios where the driver felt that either the speed while approaching or driving through the scenario was not well chosen, so they corrected it based on their personal preference.
Most of these interventions are scenarios where drivers preferred a higher speed while approaching or driving through a roundabout at \unit[32.0]{\%}, while only \unit[3.5]{\%} were speed decreases before or in roundabouts.
\unit[21.2]{\%} are lower speeds during the approach of or while driving through turns, while \unit[21.0]{\%} are wishes for higher speeds in that scenario instead.
The remaining percentages are made up of curve speed adjustments with \unit[13.9]{\%} of speed increases and \unit[6.8]{\%} of speed decreases in curves.
The remaining percentages are made up of \unit[12.9]{\%} of speed increases in curves, \unit[7.5]{\%} of speed decreases in curves and \unit[1.9]{\%} of speed decreases before give way situations on straight roads.

The adjustment of acceleration and deceleration timing onto speed limits at \unit[10.3]{\%} of the rebalanced distribution is made up of two different intervention types.
The PLDF always chooses its timing for acceleration and deceleration to comply with the legal speed limit.
This means that the PLDF will only accelerate after a higher speed limit sign is passed and preemptively decelerate to reach lower speed limits at the appropriate speed.
Some drivers however prefer a less passive driving style.
\unit[68.8]{\%} of this category is made up of earlier accelerations onto higher upcoming speed limits while later decelerations onto lower upcoming speed limits make up for \unit[31.2]{\%}.

The adjustment of acceleration strength only describes scenarios in which drivers used the gas to reach a stronger acceleration to the current speed limit.
This is most often the case after leaving turns or roundabouts.
However, at \unit[2.3]{\%} of the rebalanced distribution, this intervention is relatively rare and only occurs with a few specific drivers.

Interesting to mention is also that corrections of the ACC behavior only made up for \unit[1.4]{\%} of the rebalanced distribution.
The most common scenario in this case was pressing the gas in stop-and-go situations, especially at traffic lights, in order to decrease the distance to the leading target vehicle.
Drivers often described the ACC behavior in these scenarios as "too sluggish".
The other interventions were gas and brake pedal usage to adjust the distance to the target vehicle in other scenarios, however, these cases were rare.
However, it is important to note that this statistic does not take into account any changes made to the ACC time gap setting, which were not analyzed in this study.

\subsubsection{Interventions due to System Limitations}

Traffic interactions account for \unit[27.7]{\%} of the rebalanced distribution.
The most common traffic interactions were braking at give-way situations at \unit[47.2]{\%} and braking at red traffic lights at \unit[24.7]{\%}.
The reactivation of the PLDF after long stops is the third most common traffic-related reason at \unit[11.1]{\%}.
The remaining \unit[17.0]{\%} consist of overtaking maneuvers, lane changes, merging, reactions to cut-ins by other drivers, and others.

\subsubsection{Interventions due to Wrong Input Information}

The correction of the ACC sensing state information is the most common intervention based on wrong input data at \unit[7.0]{\%} of the rebalanced distribution.
It consists to \unit[46.6]{\%} of not detected targets and to \unit[22.3]{\%} of wrong targets, e.g., a standing car on the side of the road which is wrongly detected as a target vehicle.
\unit[24.0]{\%} are due to vehicles that are kept as targets by the PLDF but are no longer relevant, e.g., due to changing to another lane or turning.
The remaining \unit[7.2]{\%} represent target losses, which refers to vehicles that were initially detected but left the system's field of view prematurely.

The correction of incorrect traffic sign recognition and map information category at \unit[5.3]{\%} of the rebalanced distribution cannot be split into smaller categories since it is often unclear from the recorded data whether the erroneous information came from the map or the traffic sign recognition.

\subsection{Differences in Individual Drivers}
The data show clear differences in the behavior of individual drivers.
E.g., the most common intervention based on personal preference, the increase of the set speed, is performed by most drivers at least once, however, two out of 17 drivers never increased the set speed above the legal speed limit.
Conversely, the decrease of the set speed below the legal speed using the control lever is only performed by three out of 17 participants at least once.

Other common interventions include the adjustment of the acceleration and deceleration timing onto upcoming speed limits.
There, significant differences can also be observed in the individual drivers.
E.g., for the earlier acceleration onto upcoming higher speed limits, more than \unit[50]{\%} of the interventions are performed by five specific drivers.
For the later deceleration onto upcoming lower speed limits, this is even more severe where six drivers make up over \unit[80]{\%} of the cases this intervention was performed.
These specific drivers perform this intervention regularly and would benefit from an adjustment of the PLDF to incorporate later decelerations in this scenario.
For most other drivers, such an adjustment would not be suitable since they seem satisfied with the PLDF's behavior in this scenario.
These are just a few examples, but the distribution of other interventions due to personal preference also shows significant differences between individual drivers.

However, it must be stated that drivers have different commuter routes and the types of interventions that can occur for them may differ.
E.g., a participant with no sharp curves or turns on their commute cannot intervene in that scenario.
For more precise driver profiles, their behavior on the same route should be compared instead.

\begin{table*}[b]
\renewcommand{\arraystretch}{1.3}
\caption{Spearman and Kendall correlation coefficients and p-values with the driver satisfaction for selected features}
\label{tab:correlation}
\begin{center}
\begin{tabular}{cccccc}
\toprule 
\textbf{Feature} & \textbf{Spearman $\rho$} & \textbf{Spearman p-value} & \textbf{Kendall $\tau$} & \textbf{Kendall p-value} & \textbf{Significant}\tabularnewline
\midrule
\midrule 
Number of interventions per drive & -0.533 & 0.027 & -0.411 & 0.039 & yes\tabularnewline
\midrule 
Number of interventions per minute & -0.602 & 0.011 & -0.465 & 0.019 & yes\tabularnewline
\midrule 
Number of interventions within the ODD per drive & -0.569 & 0.017 & -0.450 & 0.024 & yes\tabularnewline
\midrule 
Number of interventions within the ODD per minute & -0.528 & 0.029 & -0.411 & 0.039 & yes\tabularnewline
\midrule 
Number of interventions outside the ODD per drive & -0.299 & 0.244 & -0.215 & 0.281 & no\tabularnewline
\midrule 
Number of interventions outside the ODD per minute & -0.315 & 0.219 & -0.268 & 0.177 & no\tabularnewline
\midrule 
Average duration per drive & -0.032 & 0.903 & -0.018 & 0.928 & no\tabularnewline
\midrule 
Average distance per drive & -0.121 & 0.644 & -0.089 & 0.653 & no\tabularnewline
\bottomrule
\end{tabular}
\end{center}
\end{table*}

\subsection{Correlation Analysis}
Subsection \ref{subsec:questionnaire} demonstrated that drivers subjectively desire fewer interventions, indicating a correlation between driver satisfaction and intervention behavior.
To test whether this can be confirmed objectively, a correlation analysis was conducted on selected features with the general driver satisfaction values from the questionnaire.
It was checked which features have a significant effect on driver satisfaction.
Since the data are sparse, containing only 17 participants, the Spearman correlation and Kendall correlation were chosen.
They are also applicable since most features are non-normal distributed and some features contain outliers.
Mainly interesting are the effects that the total number of interventions per drive and the intervention frequency per minute have on driver satisfaction.
In order to check whether interventions within or outside the system's ODD have a different effect on satisfaction, their numbers and frequencies were also tested.
A correlation is considered to be significant if the p-values for both correlation methods are below a significance level of 0.05.
The results are shown in Table \ref{tab:correlation}.

As can be seen, four features have a significant correlation with driver satisfaction: The number and frequency of interventions in general, but also especially of interventions within the PLDF's ODD.
Interventions outside of the PLDF's ODD do not seem to have such a significant effect on driver satisfaction.
Perhaps the drivers are aware that the system was not designed to handle such situations, and therefore, they are more lenient with it.
To determine if the number of interventions per drive is correlated to driver satisfaction and not just the duration or distance of the commute, the effect of the average duration and distance per drive was also tested.
With very low calculated correlation coefficients and very high p-values, it can be said that neither the duration nor the distance of the drives have a significant effect on driver satisfaction.
This suggests that especially the interventions within the ODD, which could be avoided by adjusting the PLDF's behavior and the quality of its input data, show a significant potential for optimization in order to increase customer satisfaction with the system.

\section{Summary and Outlook}

In this paper, it was shown that drivers perform a high number of driver-initiated takeovers during the use of a SAE level 1 PLDF.
Out of these interventions, \unit[53.7]{\%} occur due to deviating driver preferences from the intended function behavior.
This shows a high potential for ADAS behavior optimization in order to better fit the drivers' personal preferences.
These preferences can differ considerably from driver to driver, proving the need for ADAS individualization.
Using the questionnaire results and a correlation analysis, a significant negative impact of the number and frequency of driver interventions during active function use on the driver satisfaction was found.
This correlation is much stronger for interventions within the ADAS's ODD compared to interventions in situations that the function was not designed to handle.
13 out of 17 participants stated in the questionnaire that they were satisfied with the ADAS's behavior outside of the situations they intervened in.
Therefore, using these interventions could be a good starting point for ADAS individualization for most drivers.

Future work should analyze the different driver profiles and also the consistency of driver behavior over multiple drives on the same route in more depth.
Using the intervention data, analyses of the potential ways of ADAS optimization and individualization should also be conducted.
After implementing the changes necessary based on these analyses, another test group study should be conducted in order to verify whether drivers become more satisfied if the ADAS is adjusted based on their intervention behavior.
Similar studies on the takeovers in ADAS with SAE levels 2 or higher should also be conducted in the future.

\bibliography{IEEEabrv, test_group_study_references}

\end{document}